\def\year{2019}\relax
\documentclass[letterpaper]{article} 
\usepackage{aaai19}  
\usepackage{times}  
\usepackage{helvet}  
\usepackage{courier}  
\usepackage{url}  
\usepackage{graphicx}  
\usepackage{color}
\usepackage{multirow}
\usepackage{caption, afterpage}
\usepackage{amsmath,amsfonts,amsthm,amssymb}
\usepackage{svg}
\begin{document}

\title{Comparing Observation and Action Representations for \\ Deep Reinforcement Learning in $\mu$RTS}

\author{Shengyi Huang \textnormal{and} Santiago Onta\~{n}\'{o}n\\
Drexel University, Philadelphia, Pennsylvania, 19104\\
\{sh3397,so367\}@drexel.edu
}

\maketitle
\begin{abstract}

This paper presents a preliminary study comparing different observation and action space representations for Deep Reinforcement Learning (DRL) in the context of Real-time Strategy (RTS) games. Specifically, we compare two representations: (1) a {\em global} representation where the observation represents the whole game state, and the RL agent needs to choose which unit to issue actions to, and which actions to execute; and (2) a {\em local} representation where the observation is represented from the point of view of an individual unit, and the RL agent picks actions for each unit independently. We evaluate these representations in $\mu$RTS showing that the local representation seems to outperform the global representation when training agents with the task of harvesting resources.
\end{abstract}

\section{Introduction}

Real-time strategy (RTS) games pose a significant challenge for artificial intelligence (AI)~\cite{buro2003real,ontanon2013survey}. They are complex due to a variety of reasons: (1) players need to issue actions in real-time, which means agents have a very limited time to produce what is the next action to execute, (2) most RTS games are partially observable, i.e., a player might not always able to observe the opponents' strategies and actions, and (3) RTS games have very large action spaces.  

Recent application of Deep Reinforcement Learning (DRL) to RTS games introduces additional challenges such as (1) dealing with extremely sparse rewards and (2) designing efficient observation and action space representations. In this paper, we attack the latter challenge by comparing two intuitive representations. The first one is a  {\em global} representation that enables the DRL agent to observe the entire game state as a series of features, and issue global commands to choose both which unit to issue actions to and which actions to execute. The second one is a {\em local} representation where the DRL agent sequentially selects actions for individual units independently. We used $\mu$RTS as our testbed to evaluate the performance of these representations and our experiments show that: (1) the local representation seems to outperform global representation, probably because the agent using local representation does not have to learn how to select a unit; (2) both local and global representation significantly outperform a random baseline agent.


The remaining of this paper is organized as follows. In the next section, we will present some background on RTS games and reinforcement learning. Then, we present our training methods, experiments setup, and discussion of training results. Lastly, we make our conclusion and consider the future works.

\section{Background}\label{sec:background}

This section provides the necessary background on RTS games, and reinforcement learning in RTS games.

\subsection{Real Time Strategy Games}\label{subsec:microrts}

Real-time Strategy (RTS) games are complex adversarial domains, typically simulating battles between a large number of military units, that pose a significant challenge to both human and artificial intelligence \cite{buro2003real}. Designing AI techniques for RTS games is challenging because (1) they have {\em huge decision spaces}: the branching factor of a typical RTS game, StarCraft, has been estimated to be on the order of $10^{50}$ or higher \cite{ontanon2013survey}; (2) they are {\em real-time}, which means that these games typically execute at 10 to 50 decision cycles per second, leaving players with just a fraction of a second to decide the next action, players can issue actions simultaneously, and actions are durative, and (3) most of them are partially observable and players must sent units to scout the map.

\begin{figure}[t]
    \centering
    \includegraphics[width=0.7\columnwidth]{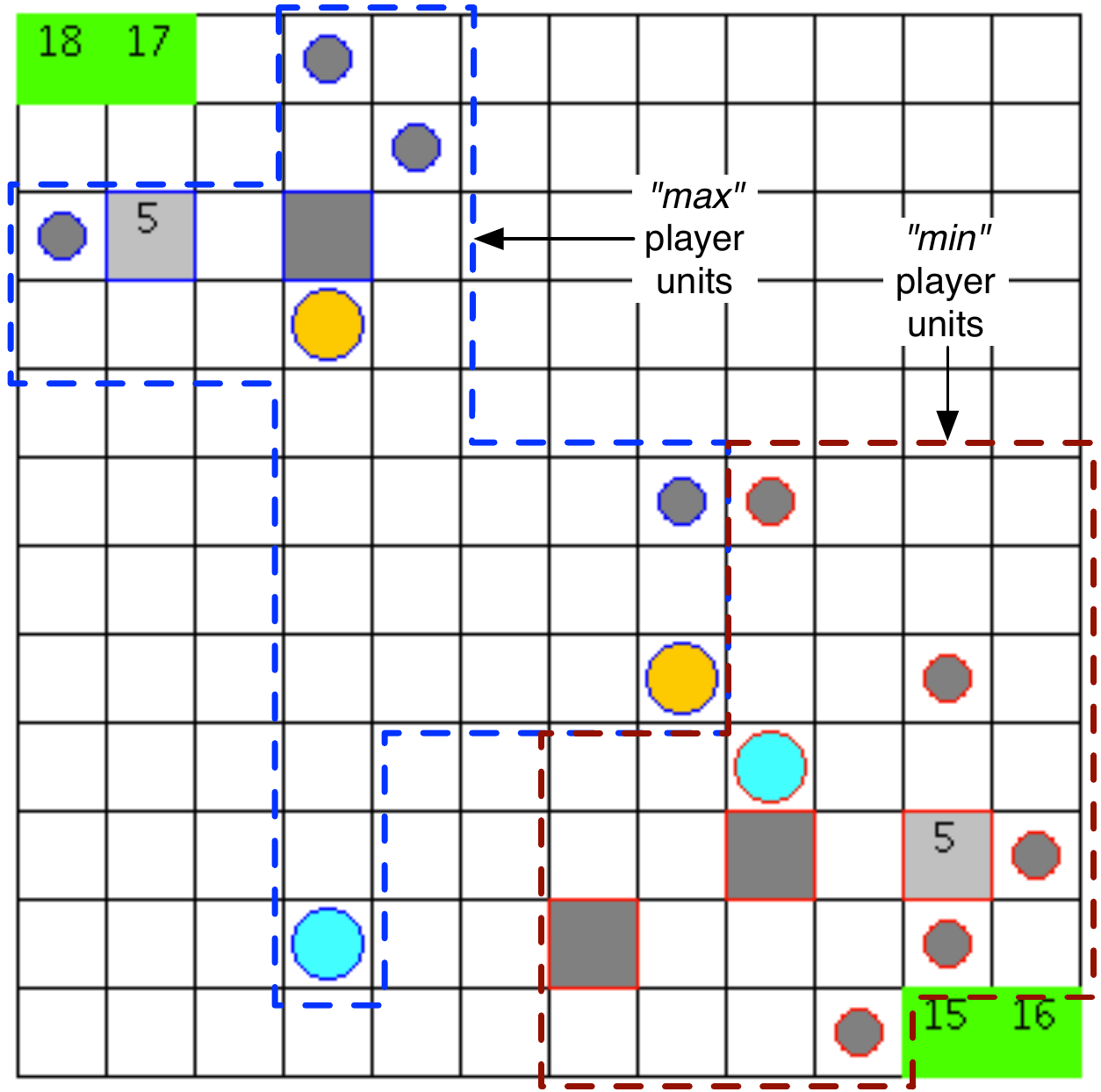}
    \caption{A screenshot of $\mu$RTS. Square units are ``bases'' (light grey, that can produce workers), ``barracks'' (dark grey, that can produce military units), and ``resources mines'' (green, from where workers can extract resources to produce more units), the circular units are ``workers'' (small, dark grey) and military units (large, yellow or light blue).}
    \label{fig:microRTS}
\end{figure}

In the experiments reported in this paper, we employed $\mu$RTS\footnote{\url{https://github.com/santiontanon/microrts}}, a simple RTS game maintaining the essential features that make RTS
games challenging from an AI point of view: simultaneous and durative actions, large branching factors and real-time decision making. Although the game can be configured to be partially observably and non-deterministic, those settings are turned off for all the experiments presented in this paper. 

A screenshot of the game is shown in Figure~\ref{fig:microRTS}. The squared units in green are Minerals with numbers on them indicating the remaining resources. The units with blue outline belong to player 1 and those with red outline belong to player 2. The light grey squared units are Bases with numbers indicating the amount of resources owned by the player, while the darker grey squared units are the Barracks. Movable units have round shapes with grey units being Workers, orange units being Lights, yellow being Heavy units (not shown in the figure) and blue units being Ranged. 

Moreover, in order to ease reinforcement learning research in $\mu$RTS, we have prepared an OpenAI gym wrapper for the game, which we have made available to the research community\footnote{\url{https://github.com/vwxyzjn/gym-microrts}}.

\subsection{Reinforcement Learning in RTS Games}\label{subsec:crl}

One of the earliest applications of reinforcement learning to RTS games is the work of Concurrent Hierarchical Reinforcement Learning (CHRL) with Q-learning \cite{marthi2005concurrent}. Marthi et al. experimented with the Wargus game and demonstrated the use of Alisp language to specify a list of desired tasks and use Q-learning to tune the parameters. They showed that the CHRL could significantly improve the agents' performance compared to flat concurrent hierarchical Q-learning.

In addition, Jaidee and Mu{\~n}oz-Avila have leveraged Q-learning  to learn a policy for each class of units (peasants, knights, barracks, etc), which drastically reduced the memory requirement of Q-learning \cite{modeling}. The trade-off is the lack of coordination between agents since each agent only learns to control a certain type of units or buildings. That being said, they were able to demonstrate good performance by defeating the built-in AI 80\% of the time.

Another idea explored in RTS games is that of {\em options}, which are temporally extended actions \cite{stolle2002learning}. The agent first needs to evaluate which option to choose, and then needs to determine when to terminate the options chosen. The options framework also exhibits a sense of Hierarchical Reinforcement Learning (HRL) since raw actions are primitives compared to options. Researchers have tried to combine options and heuristic algorithms to simplify the game space \cite{Tavares2018TabularRL}.

In recent years, {\em Deep Reinforcement Learning (DRL)} approaches have received significant attention in a number of games such as classic Atari games \cite{mnih2013playing} and Go \cite{silver2017mastering}. Over the years, researchers have introduced different type of DRL algorithms. Most notably, Asynchronous Advantage Actor-Critic (A3C), Deep Q-Network (DQN), and Proximal Policy Gradient (PPO) have gained the state-of-the-art results in a variety of games \cite{mnih2016asynchronous,mnih2013playing,schulman2017proximal}.

In 2017, researchers at Deepmind started tackling the challenge of training agents for StarCraft II, one of the most complex and popular RTS games in history \cite{vinyals2017starcraft}. They conducted a series of experiments using A3C \cite{mnih2016asynchronous} on the full game as well as a collection of mini-games such as training the ``marines'' units to defeat the ``roaches'' units in StarCraft II. They were able to show good converging performance for the mini-games but had no material success in the full game setting. However, they recently demonstrated a bot ``AlphaStar'' that defeats a professional StarCraft II Protoss player in the full game setting.

Researchers at Facebook also joined this line of research by publishing an open-sourced RTS game research platform ELF \cite{tian2017elf}. 
For simplicity, they conducted experiments utilizing hard-coded actions such as BUILD\_BARRACKS, which ``automatically picks a worker to build barracks at an empty location, if the player can afford''. Through these high-level actions, they trained agents using A3C, curriculum training, MCTS, and show the trained agent is able to defeat the built-in scripted bots in the full-game setting. Notably, researchers at Tencent used similar approach to hard-code high-level actions in StarCraft II and defeated the cheating AI in the full-game setting by commencing early attacks \cite{sun2018tstarbots}. Compared to this work, we are interested in RL settings that do not require hand-crafted macro actions.

Moreover, Liu et al. studied the application of HRL to incorporate the generation of macro actions through expert replays \cite{liu2019efficient}. Lee et al. also hard-coded a collection of macro actions for zergs and trained agents combining DQN and LSTM, obtaining a 83\% winning rate in the AIIDE 2017 StarCraft AI competition \cite{lee2018modular}.

To address the action coordination problem, Samvelyan, Rashid et al. built the StarCraft Multi-Agent Challenge (SMAC) for Multi-agent Reinforcement Learning (MARL), where each unit is controlled by a separate agent~\cite{samvelyan2019starcraft}. They compared popular MARL algorithms such as QMIX, COMA, VDN, and IQL in mini maps that usually features battle between different agents \cite{rashid2018qmix,foerster2018counterfactual,sunehag2017value,iql}.

Though these previous research show important results, there has not been any published study comparing the relative merits of different types of observation and action representations in RTS games, which is the key issue that our paper starts to address.

\begin{table}[t]
\centering
\begin{small}
\begin{tabular}{ |c|c|c|c| } 
\hline
\multicolumn{2}{|c|}{\textbf{Global Representation}} \\
\hline
Features  & Possible values \\
\hline
Hit Points & 0, 1, 2, 3, 4, 5, $\geq 6$  \\ 
Resources& 0, 1, 2, 3, 4, 5, $\geq 6$  \\ 
Owner& player 1, -, player 2 
\\ 
Unit Types & -, resource, base, barrack, worker, light, heavy \\ 
Action& -, move, harvest, return, produce, attack\\ 
\hline
\end{tabular}

\vspace*{0.5cm}
\begin{tabular}{ |c|c|c|c| } 
\hline
\multicolumn{2}{|c|}{\textbf{Local Representation}} \\
\hline
Features  & Possible values \\
\hline
Hit Points & wall, 0, 1, 2, 3, 4, 5, $\geq 6$  \\ 
Resources& wall, 0, 1, 2, 3, 4, 5, $\geq 6$  \\ 
Owners& wall, player 1, -, player 2 
\\ 
Unit Types& wall, -, resource, base, barrack, worker, light, heavy  \\ 
Action& wall, -, move, harvest, return, produce, attack \\ 
\hline
\end{tabular}
\end{small}
\caption{The list of feature maps and their descriptions.}
\label{tab:features}
\end{table}

\section{Our Approach}\label{sec:representations}

We designed two pairs of observation and action representation for comparing performance. The first one is a  global representation that enables the RL agent to observe the entire game and issue global commands to choose both which unit to issue actions to and which actions to execute. The second one is a local representation where the agent is externally given a specific unit to control and just has to pick actions for that unit. Both representations are designed to train agents to harvest resources as fast as possible. 

\subsection{Global Representation}\label{subsec:micrortsglobalagent}

We utilized \emph{flattened one-hot} feature maps as our observation representation similar to PySC2 \cite{vinyals2017starcraft} and a couple pre-existing works on observation evaluation functions on $\mu$RTS \cite{inproceedings,Yang2018LearningME}. Assuming the game map is represented as a grid, a feature map is a matrix of the same dimensions as the game map, where in each cell of the matrix we have a value that represents some aspect of the corresponding game map grid cell. For simplicity, we assume that each feature map takes discrete values. The exhaustive list of feature maps and their description is presented at Table \ref{tab:features}.

Let $h$ and $w$ be the height and width of the map respectively, we construct the {\em observation} as a set of $n_f=5$ feature maps and that all feature maps can take the same number of values $n_c$. Thus, each feature map is a vector of length $h * w$, and the observation can be represented as a $n_f \times (h*w)$ matrix. Finally, in order to facilitate learning, we use a one-hot-encoding representation, and thus the observation is represented as a $n_f \times (h*w) \times n_c$ tensor ($n_c$ is the number of values of the feature plane that can take the larger number of values). The dimensions of these tensors are usually referred to as their {\em shape} in common deep learning libraries, and thus we will use this term from now on. Moreover, in the experiments reported in this paper $n_f = 5$ and $n_c = 7$.

Then, to execute an action in $\mu$RTS, the RL agent needs to specify the unit, the action types, and the parameters of the action to execute. For our experiments that only focus on the task of harvesting resources, the set of target action types are NOOP (no operation, represented simply as a - in Table \ref{tab:features}), Move, Harvest, Return. Some of these action types requires an action parameter that specifies the direction at which the action is issued. The list of available action parameter is Up, Right, Down, Left. So if the agent wants to command a worker to harvest the resources located left to it, the agent has to specify the worker, set action type to Harvest, and action parameter to Left.

Since unit IDs change from game to game, in order to learn generalizable policies, the global representation does not require the agent to specify the unit ID. Instead, the RL agent predicts the coordinates of the unit it wants to control.

Therefore, we construct the \emph{action} at time step $t$ as a vector with 4 elements: $\left[a_t^{\text{x}},  a_t^{\text{y}}, a_t^{\text{action type}}, a_t^{\text{action param}}\right]$,
where $a_t^{\text{x}}, a_t^{\text{y}}, a_t^{\text{action type}} , a_t^{\text{action param}}$ are integers signifying the selected x-coordinate, y-coordinate, action type, and action parameters, respectively, predicted using one-hot encoding representations. Moreover: 
\begin{enumerate}
    \item If the action produced is invalid in $\mu$RTS, the action will be replaced by a NOOP action. Thus, it is up to the RL agent to learn which are the valid actions.
    \item This representation varies from PYSC2's action representation that uses the actions themselves to select units and perform actions on them.
    \item The agent can only issue one action to one unit at each time step. 
\end{enumerate}

In summary, the global representation encodes the entire game state for the RL agent, who needs to learn to choose which unit to control (by outputting the unit's coordinates) and which action type and parameter to be issued at each time step.

\begin{figure}[t]
  \centering
  \includegraphics[width=0.8\columnwidth]{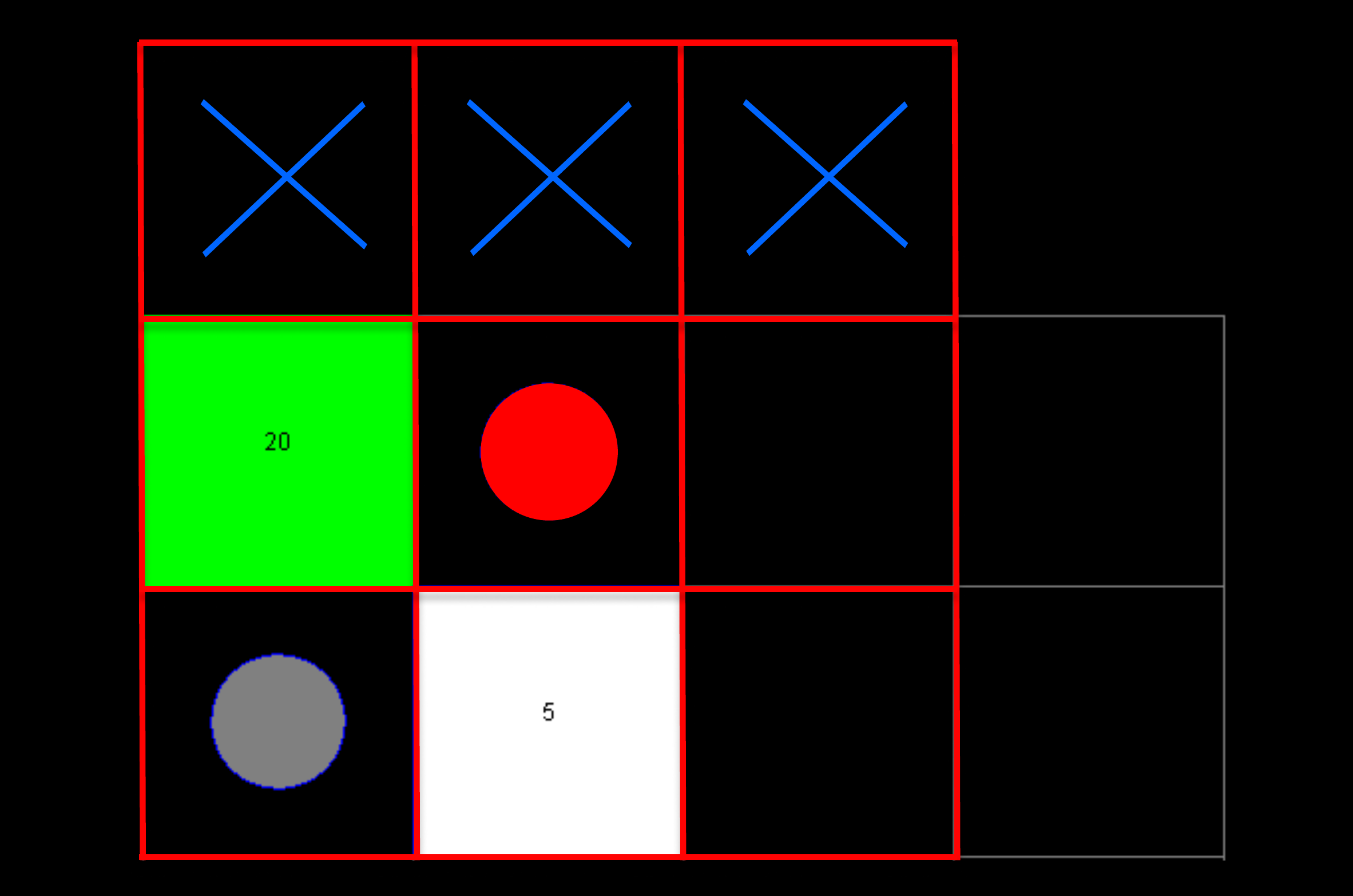}
  \caption{The local feature maps of shape $(2w+1) \times (2w+1) = 3 \times 3$ outlined in red of the an unit (the red circle) when $w=1$. The blue crosses indicate the cells are the walls of the game map.}
  \label{fig:local}
\end{figure}

\subsection{Local Representation}\label{subsec:micrortsglobalagent}
In this representation, the agent perceives the game from the point of view of a particular unit and only needs to issue action to that unit. In other words, the agent does \emph{not} have to learn to pick which unit to control. 

We utilized flattened one-hot \emph{local} feature maps as our observation representation. Consider a parameter of window size $w$ that specifies ``how far can an unit see''. Specifically, a window size $w=1$ means each unit is able to observe the cell it is located and all the cells that are 1 distance away from it in either axis, as shown in \ref{fig:local}. Note that if the unit sees a cell beyond the boundary of the map, we consider that cell a ``Wall''  (Note that the global representation does not need such ``Wall'' unit type because the agent always sees the entire map and does \emph{not} see beyond the boundary of the map). A local feature map is thus a matrix of dimension $(2w +1)\times(2w +1)$. A comprehensive list of local feature maps and their descriptions can be found in Table \ref{tab:features}. 

At each time step, the agent is given a target unit (we rotate which unit is the focus at consecutive game frames). For instance, assuming there are 3 units $u_1, u_2, u_3$, then $u_1, u_2, u_3, u_1, u_2, ... $ will be selected at time step $1, 2, 3, 4, 5, ...$, respectively. Given the selected unit at each time step, we construct \emph{observation} to be a set of $n_f$ local feature maps from the point of view of the said unit according to $w$, with all of them can take the same number of discrete values $n_c$. Similarly, we use one-hot-encoding representation and the observation is represented as a $n_f \times (2w+1)^2 \times n_c$ tensor, where $n_f = 5, n_c=8$.

For action execution, since the unit is already selected, the agent only needs to predict the action type and parameter. Therefore, we construct the \emph{action} at time step $t$ as a vector with 2 element: $\left[a_t^{\text{action type}}, a_t^{\text{action param}}\right]$, 
where $a_t^{\text{action type}} , a_t^{\text{action param}}$ are integers that signifies the selected action type, and action parameters, respectively (predicted as one-hot vectors). 
As before, if the action issued is not valid in $\mu$RTS, it will be replaced by a NOOP action.

In summary, the local representation contains the local game state features from the point of view of a selected unit for the RL agent, and it needs to learn which action type and parameter to be issued at each time step. Note that it does \emph{not} have to learn how to select a unit to control, but only how to interact with its surrounding observations.

\begin{figure*}[t]
  \centering
  \includegraphics[width=0.8\textwidth]{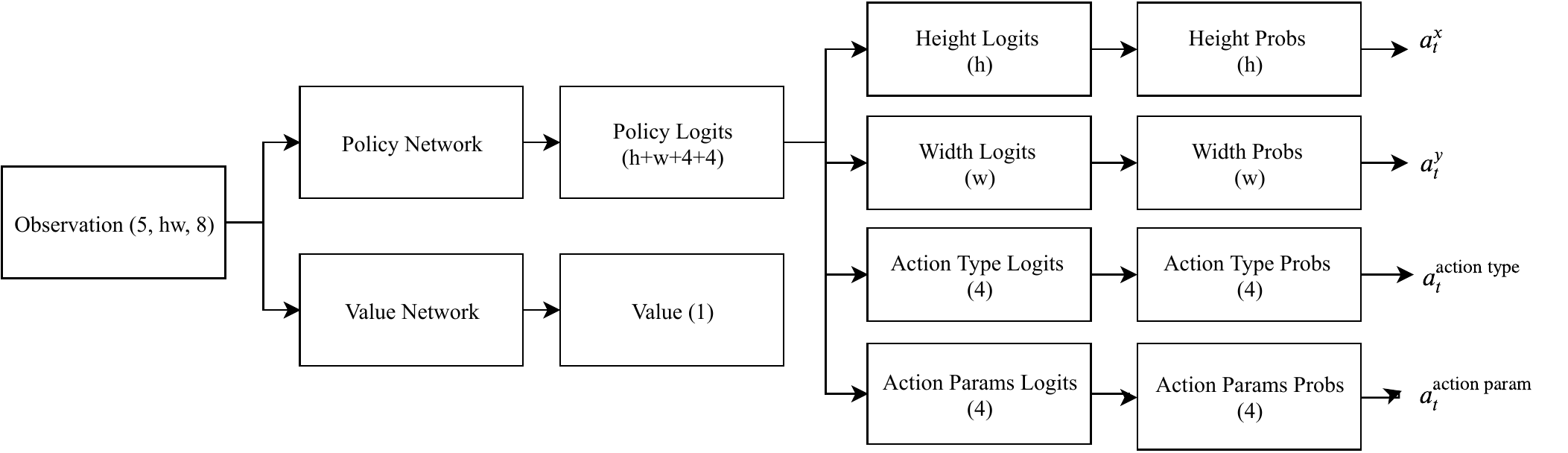}
  \caption{The neural network architecture that demonstrates the flow from the observation vector to action probabilities. The number at each box suggests the input or output shapes  }
  \label{fig:training}
\end{figure*}

\subsection{Reward Function}\label{subsec:reward}

The game environment will give the agent a reward of 10 when a worker has successfully harvested resources and another 10 when it returns  the  harvested resources back to the base. Otherwise the game environment gives a reward of 0. 

It's worth pointing out that, like in many other reinforcement learning problems, rewards can be very sparse as maps grow and workers get further away from the resources and the base, requiring thousands of exploration steps before getting a non-zero reward.


\section{Training of Agents}\label{sec:training}

\begin{figure}[tb]
  \centering
  \includegraphics[width=\columnwidth]{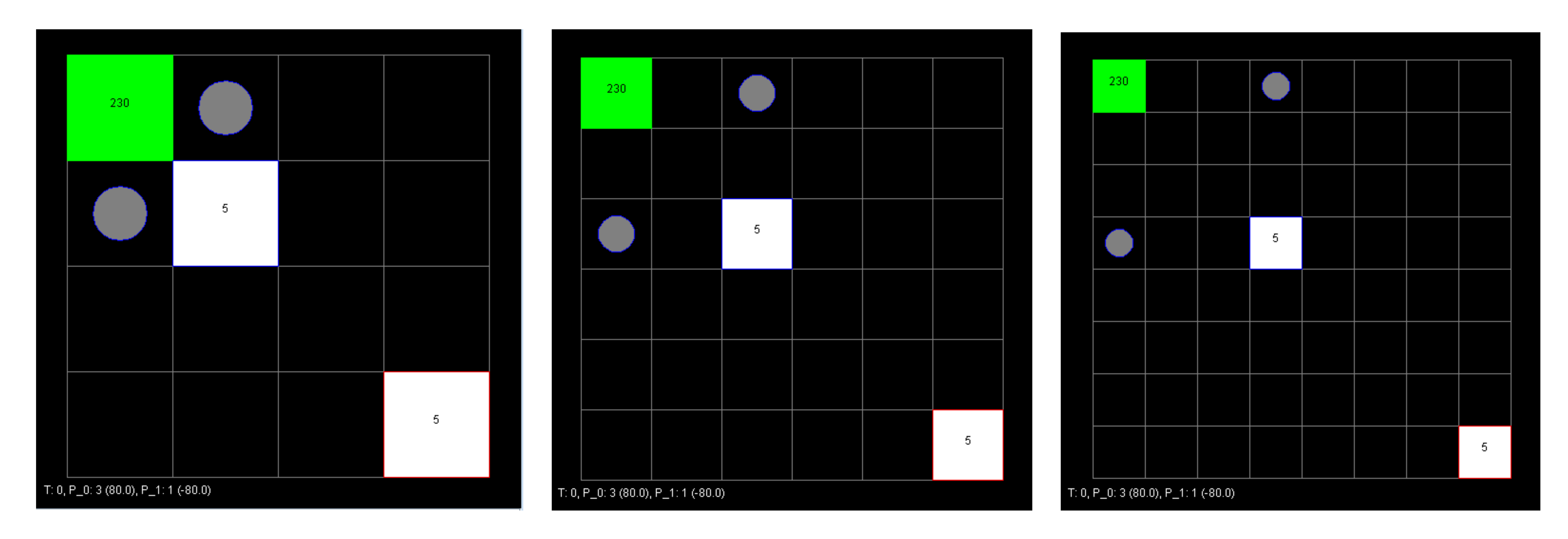}
  \caption{The mini-games that focuses on harvesting resources with different map sizes of $4\times4$, $6\times6$, and $8\times8$. 
  }
  \label{fig:games}
\end{figure}

For the global representation, we selected Advantage Actor-Critic (A2C), one of the most popular RL algorithms, to evaluate the agents' performance \cite{mnih2016asynchronous}. 
We created the Policy Network and Value Networks. As shown in Figure \ref{fig:training} that demonstrates the forward pass, the observation tensor is fed into the policy network that produces a dense vector of shape $(h\cdot w \cdot n_a \cdot n_p)$, which is split into 4 different subvectors of logits. Each of those logits is passed into a Categorical Distribution to gain the probability distribution for sampling values to produce the a combined action vector $a=\begin{bmatrix}
         a_t^{\text{x}} &   a_t^{\text{y}} & a_t^{\text{action type}}  & a_t^{\text{action param}} 
        \end{bmatrix}$. In addition, the observation vector is also used by the value network to create a valuation of the current observation.

\begin{table}[t]
\centering
\begin{small}
\begin{tabular}{ |l|l|l|c| } 
\hline
Parameter Names  & Parameter Values\\
\hline
Maps & $4\times4$, $6\times6$, $8\times8$ \\ 
Learning Rate of Adam Optimizer&  0.0007  \\ 
Random Seed & 1, 2, 3  \\ 
Episode Length& 2,000 time steps  \\ 
Total Time steps & 2,000,000 time steps \\ 
$w$ (Local Representation Window Size) & 1, 2 \\
$\gamma$ (Discount Factor) & 0.99 \\ 
$\beta$ (Value Function Coefficient) & 0.25 \\ 
$\eta$ (Entropy Regularization Coefficient) & 0.01 \\ 
$\omega$ (Gradient Norm Threshold )& 0.5 \\
\hline
\end{tabular}
\end{small}
\caption{The list of experiment parameters and their values.}
\label{tab:params}
\end{table}

Given the forward pass, we run the game for a full episode, record the reward $r_t$ at each time step $t$ and compute
\begin{align*}
    \log \pi_\theta (a_t |s_t) &= \log \pi_\theta (a_t^{\text{x}} |s_t) \\
    & + \log \pi_\theta (a_t^{\text{y}} |s_t) \\
    & + \log \pi_\theta (a_t^{\text{action type}} |s_t) \\ 
    & + \log \pi_\theta (a_t^{\text{action param}} |s_t)
\end{align*}
where $s_t$ is the observation tensor at time step $t$. In addition, analogously compute the entropy $\sum_{a} \pi_\theta (a|s_t) \log \pi_\theta (a|s_t)$  as the sum of entropy for $a_t^{\text{x}}, a_t^{\text{y}}, a_t^{\text{action type}}$ and $a_t^{\text{action param}}$.
After each episode finishes, we could train the agent based on the following gradient
\begin{align*}
    & \underbrace{(G_t - v_{\theta'}(s_t)) \nabla_\theta \log \pi_\theta (a_t |s_t) }_\text{policy gradient} + \\ &
    \underbrace{\beta (G_t - v_{\theta'} (s_t)) \nabla_{\theta'} v_{\theta'}(s_t) }_\text{value estimation gradient} + \\
    &\eta \underbrace{\sum_{a} \pi_\theta (a|s_t) \log \pi_\theta (a|s_t) }_\text{entropy regularization}
\end{align*}
where $\theta'$ and $\theta$ are weights for the value network and the policy network, respectively, $\beta$ and $\eta$ are hyperparameters for controlling the contribution of value gradients and entropy regularization gradients, and $G_t$ is the discounted rewards from step $t$. 
A comprehensive list of training parameters and their values can be found at Table \ref{tab:params}. For more details on the algorithm, please refer to the original paper \cite{mnih2016asynchronous}. 

Regarding the local representation, we used a similar Policy Network architecture that produces a dense vector of shape $(n_a \cdot n_p)$. We similarly calculate 
\[\log \pi_\theta (a_t |s_t) = \log \pi_\theta (a_t^{\text{action type}} |s_t) + \log \pi_\theta (a_t^{\text{action param}} |s_t)\] 
and the entropy $\sum_{a} \pi_\theta (a|s_t) \log \pi_\theta (a|s_t)$  as the sum of entropy for $a_t^{\text{action type}}$ and $a_t^{\text{action param}}$; then use A2C to train the agent.

\section{Experimental Evaluation}\label{sec:experiments}

In this section, we compare the aforementioned approaches on maps of different sizes.

\subsection{Experimental Setup}\label{subsec:experimental-setup}

We created three maps of size $4\times4$, $6\times6$, and $8\times8$, all of which are shown in the Figure \ref{fig:games}. They all have two workers, a block containing 230 resources, a base for the workers to return the harvested resources, and a dummy enemy base that is useless. Note that it takes a worker 10 time steps to harvest a resource and another 10 time steps to return the resources to the base, excluding the time steps to move around the map. So in the most efficient harvesting setup as shown in the $4\times4$ map, where each worker is right next to the base and the resources, the most resources that the two workers can gather is 2 per 20 frames. In fact, since both representations only accept one action at each time step, the two workers have to wait until 21 frames for the resources to be returned. Simple math shows that $2 \cdot 2000/20 = 200$ is the most that the two workers can harvest within the 2,000 time steps per episode. 
A comprehensive list of experimental parameters and their values are presented in Table \ref{tab:params}. For values of hyperparameters, we simply used the default values from the OpenAI's A2C implementation \cite{baselines}.

\subsection{Experimental Results}\label{subsec:results}

\begin{table}[tb]
\centering
\begin{small}
\begin{tabular}{ |l|l|l|l|l|  } 
\hline
    & map &  $t_{\text{first harvest}}$  & $t_{\text{first return}}$   &$r$ \\
\hline
RandomAI             &   $4 \times 4$  &  51.33      & 142.67   & 11.87     \\
Global             &   $4 \times 4$  &  99.00      & 167.73   & 13.13     \\
Local  ($w=1$)  &   $4 \times 4$  &  29.87      & 172.47   & 67.20     \\
Local ($w=2$)  &   $4 \times 4$  &  45.00      & 73.73    & 33.40     \\ \hline
RandomAI             &   $6 \times 6$  &  421.33     & 797.33   & 2.00      \\
Global             &   $6 \times 6$  &  533.33     & 1931.20  & 0.07      \\
Local ($w=1$)  &   $6 \times 6$  &  59.20      & 567.40   & 3.53      \\
Local ($w=2$)  &   $6 \times 6$  &  62.33      & 408.73   & 3.93      \\ \hline
RandomAI              &   $8 \times 8$  &  878.67     & 1480.67  & 0.87      \\
Global             &   $8 \times 8$  &  1464.53    & -  & 0.00      \\
Local ($w=1$)  &   $8 \times 8$  &  167.20     & 1844.20  & 0.20      \\
Local ($w=2$)  &   $8 \times 8$  &  89.87      & -  & 0.00      \\   
\hline
\end{tabular}
\end{small}

\caption{The list of representations and their performance according to our metrics. The ``-'' in $t_{\text{first return}}$ indicates the agent never returned any resources.}
\label{tab:metrics}
\end{table}

\begin{figure}[tb]
  \centering
  \includegraphics[width=\columnwidth]{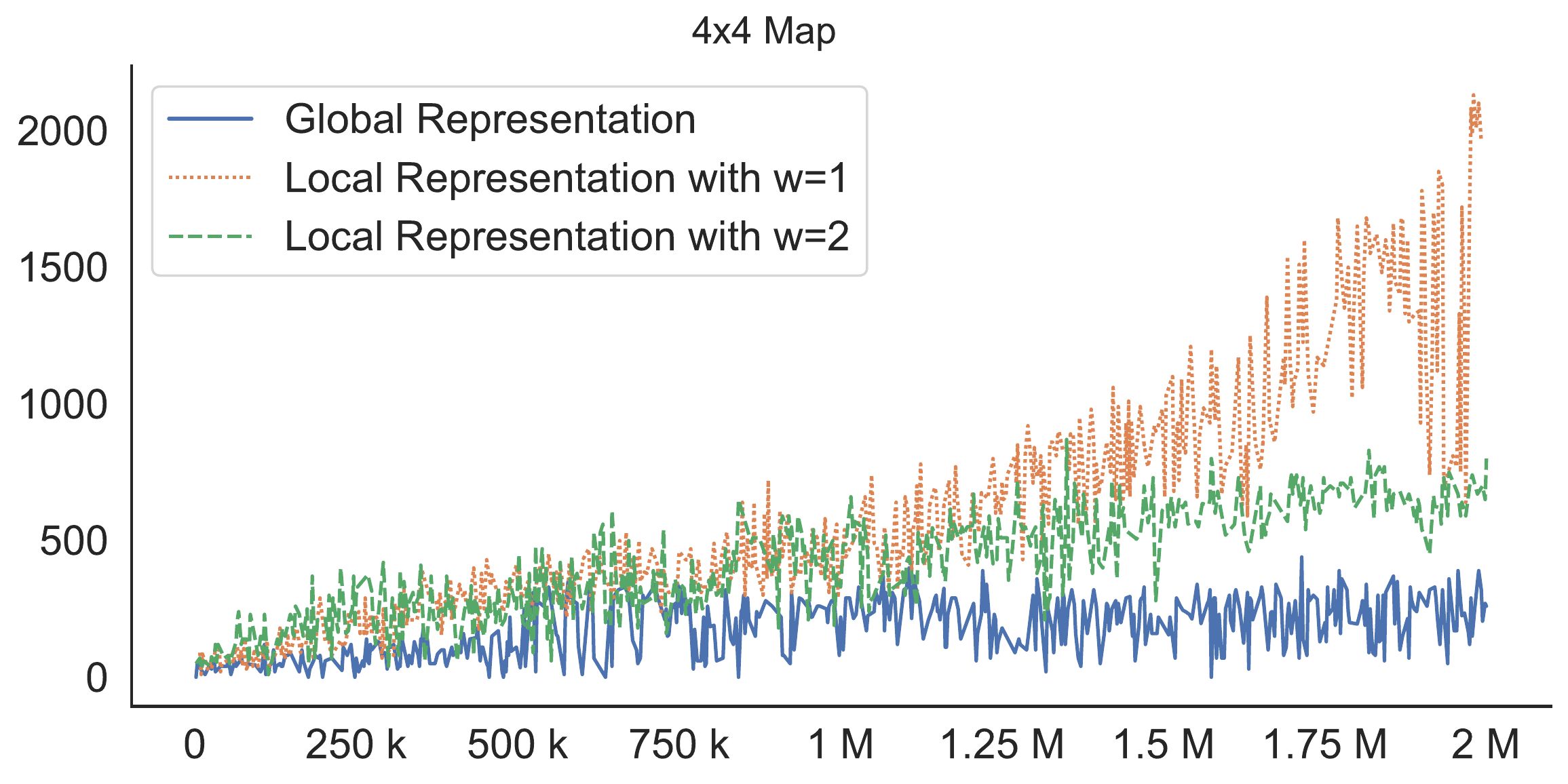}
    \includegraphics[width=\columnwidth]{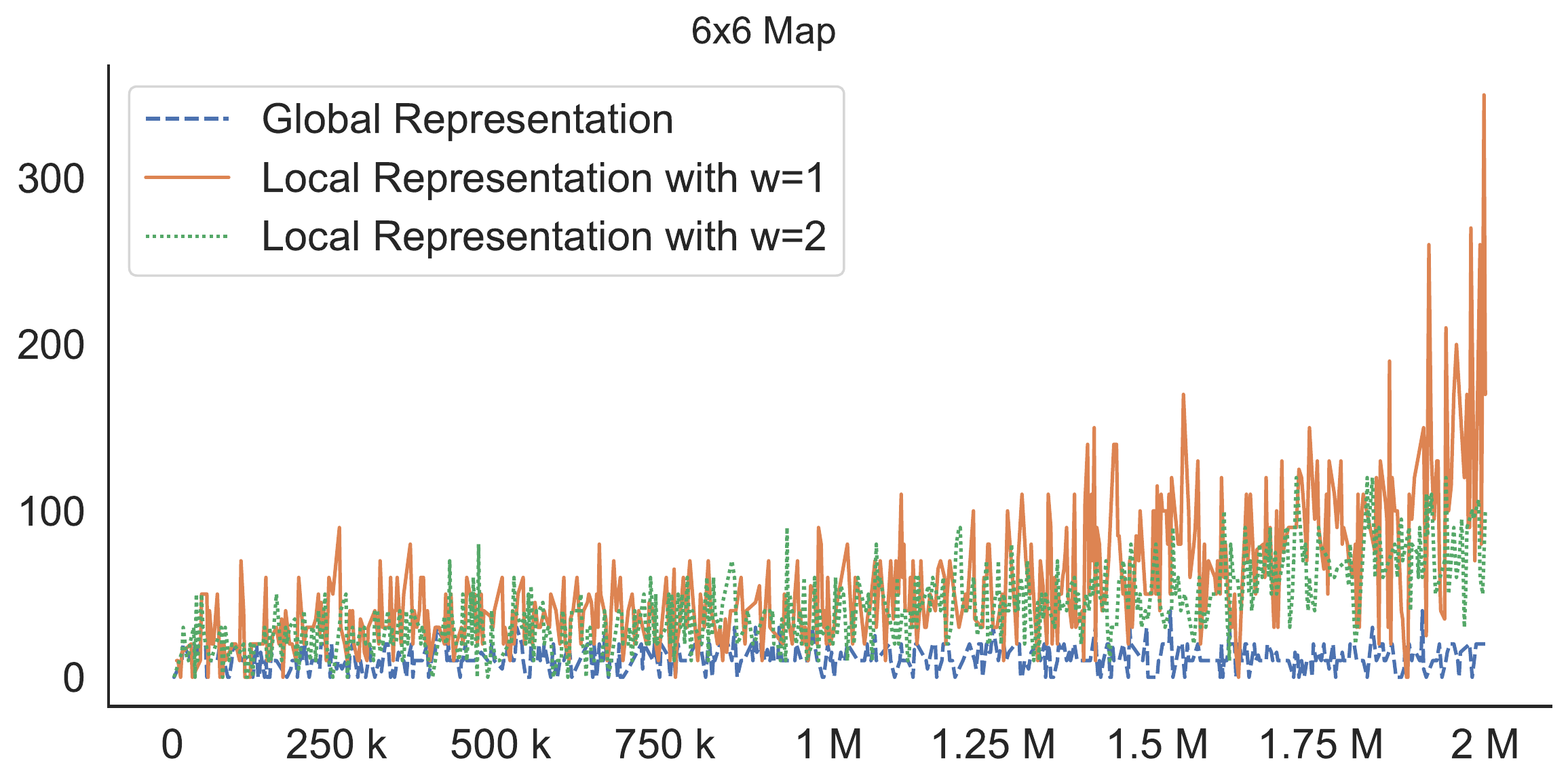}
    \includegraphics[width=\columnwidth]{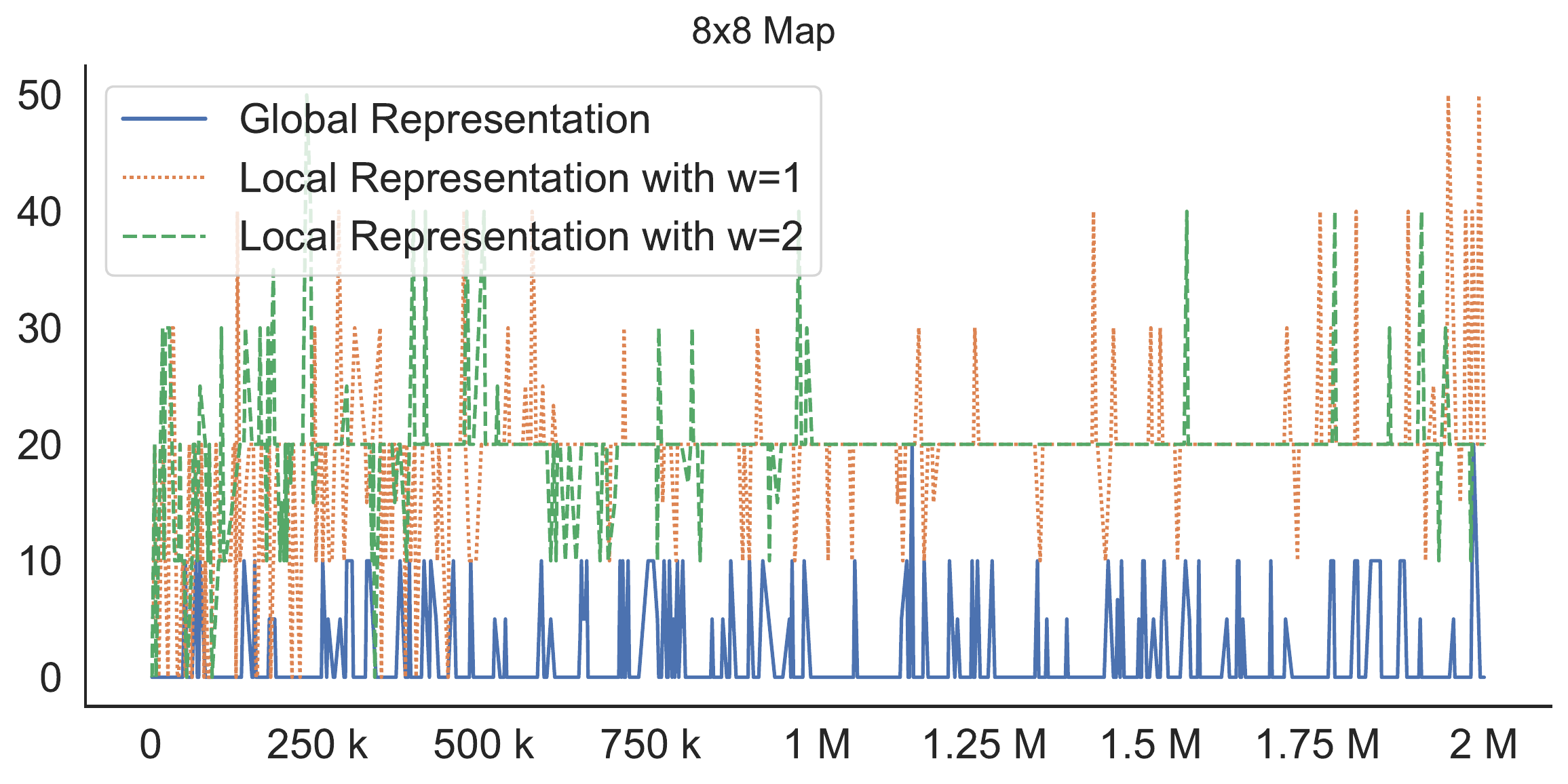}
  \caption{Episode rewards (y axis) as a function of training time steps (x-axis) for the 3 map sizes.}
  \label{fig:training_results}
\end{figure}

Figure \ref{fig:training_results} shows the average reward of runs with different random seeds per episode as a function of training steps. As shown in the plots, local representation with $w=1$ yields the best episode rewards in all three maps. As mentioned previously, we suspect local representation is performing better just because the agent doesn't have to learn to pick a unit as the agent does under the global representation, i.e., the units are pre-selected under the local representation.

The reason that local representation with $w=2$ is performing worse than local representation with $w=1$ seems purely computational since the observation tensor when $w=2$ is almost twice as the size when $w=1$. The other notable characteristic is that local representation with $w=2$ seems to exhibit more stable performance. Given more computational resources, we think local representation with $w=2$ will eventually become more efficient than local representation with $w=1$ because larger window size means the units are less likely to move to a cell where the base or resources are outside of its local feature maps (or in simpler terms, the unit moved to a cell that could not observe the base or resources). In our experiments, the map sizes are generally small and the position of the base and resources are stationary, so the local representation with $w=1$ is probably enough for the unit to observe sufficient information (and for learning to reach the resources, they just need to learn to move to the top-left coordinate of the map). 

Notably, the performance of both local and global representation suffer when we scale the game with larger maps. As shown in the Figure \ref{fig:games}, the larger maps presents even challenging moving trajectories for the agents since the distance between resources and the base is increased, which means sparser rewards and larger exploration needed for the agents. Global representation is especially worse off because the agent still has to learn to select units, navigate, harvest, and return to base in the larger maps.

As a baseline, we created three metrics to evaluate the performance of agents against the built-in RandomAI $\mu$RTS agent that uniformly selects a valid action, either NOOP, Move, Harvest, or Return, and a valid parameter depending on the game state. Notice that all actions produced by the RandomAI agent are valid since it has the game rules hard-coded, whereas our reinforcement learning agents have to learn which actions are valid. Thus, the RandomAI agent has an unfair advantage. A way to level the playing field would be to mask out invalid agents in the output of the reinforcement learning, and consider the action with highest likelihood only among the valid ones. However, we decided against giving our agent any domain knowledge beyond the observation and action representation in this first set of experiments, just to set a baseline.

Let $t_{\text{first harvest}}$ and $t_{\text{first return}}$ be the time steps that the agent successfully harvested and return resources for the first time, respectively, which will give us an idea of how much "wandering" did the agent did before harvesting and returning.  Then, let $r$ be the number of resources gathered in total, which will evaluate the agents' performance as a whole. We ran the trained agents with different seeds for 5 episodes (10,000 time steps) and calculate the average $t_{\text{first harvest}}$, $t_{\text{first return}}$, and $r$ across runs with different seeds. The evaluation result is listed at Table \ref{tab:metrics}. As shown, the trained agents perform significantly better than the RandomAI agent except in the $8 \times 8$ map where all agents perform equally bad.

\subsection{Visual Behavior of Agents}\label{subsec:behavior}

When using RL to train agents to play games, it is easy to be distracted by the numerical rewards and various metrics. However, watching the agent play in the game remains the ultimate measure for agents' ability. The following paragraphs highlight some interesting behaviors of agents.

In $4\times 4$ maps, the agent under the local representation almost exhibits optimal behavior where the two Workers harvest the resources and return them restlessly. The agent under the global representation, however, seems to only learn to control \emph{one} Worker. It seems the agent gets stuck in a local minimum where it believes the maneuver of only one Worker is best for harvesting resources.

In $6 \times 6$ and $8 \times 8$ maps, the agent under the local representation struggles with returning the resources to the base. In fact, the agent tries to harvest another unit of resources even though it already carries one.  The unit moves closer to the resources than to the base, but unfortunately it has to return the resources before harvesting new ones. It almost seems that the agent learned only to harvest resources, but has not learned how to return them.

\section{Conclusions and Future Work}\label{sec:conclusions}

In this paper, we compared two different observation and action representations in the context of RTS games: (1) a global representation that feeds the agent an observation of feature maps that are as large as the game map itself, and require the agent to learn to locate unit it is wants to issue actions to and predict the intended action type and parameter; and (2) a local representation that feeds the agent an observation of local feature maps that are the feature maps of some distance away from the point of view of a selected unit, and require the agent to learn which action type and parameter to issue for the said unit. We train agents on maps that focus on harvesting resources, establish some objective metrics to evaluate the agents' performance and show that the local representation generally outperforms the global representation. This advantage, however, does not necessarily hold in larger maps, where the exploration and sparse rewards become a huge problem.

For future works, we plan to consider full-game setting, including attack and produce actions, which have certain conditioning logic involved, i.e., the base can only produce Workers while the barracks can only produce Light, Heavy, and Ranged units. In addition, we hope to find approaches that scale better with larger map sizes. Techniques such as imitation learning or using curiosity as intrinsic rewards have been shown to help with exploring more of the map \cite{pathak2017curiosity}. Moreover, we would like to assess whether our representation comparison results generalize to other reinforcement learning algorithms.

\bibliography{references}
\bibliographystyle{aaai}

\end{document}